\documentclass[10pt,twocolumn,letterpaper]{article}

\usepackage{cvpr}              % To produce the CAMERA-READY version
\usepackage{graphicx}
\usepackage{amsmath}
\usepackage{amssymb}
\usepackage{booktabs}

\usepackage{multirow}
\usepackage{kotex}
\usepackage{xcolor}
\usepackage{ltablex}

\newcommand{\gy}[1]{\textcolor{purple}{#1}}

\newcommand{\Fref}[1]{Figure \ref{#1}}
\newcommand{\Sref}[1]{Section \ref{#1}}
\newcommand{\Tref}[1]{Table \ref{#1}}

\newcolumntype{x}[1]{>{\centering\arraybackslash\hspace{0pt}}p{#1}}
\newcommand{\specialcell}[2][c]{%
  \begin{tabular}[#1]{@{}c@{}}#2\end{tabular}}

\usepackage[pagebackref,breaklinks,colorlinks]{hyperref}

\usepackage{dsfont}

\usepackage[capitalize]{cleveref}
\crefname{section}{Sec.}{Secs.}
\Crefname{section}{Section}{Sections}
\Crefname{table}{Table}{Tables}
\crefname{table}{Tab.}{Tabs.}

\def\cvprPaperID{41} % *** Enter the CVPR Paper ID here
\def\confName{AICC}
\def\confYear{2022}

\begin{document}

\title{Memory Efficient Patch-based Training for INR-based GANs}

\author{Namwoo Lee\textsuperscript{1}\textsuperscript{2}\thanks{This work was done during an internship at NAVER AI Lab.} \thinspace\quad Hyunsu Kim\textsuperscript{1} \thinspace\quad Gayoung Lee\textsuperscript{1} \thinspace\quad Sungjoo Yoo\textsuperscript{2}
\thinspace\quad Yunjey Choi\textsuperscript{1} \\
\\
\textsuperscript{1}NAVER AI Lab \quad  \textsuperscript{2}Seoul National University
\\
% {\tt\small \{twice154}@snu.ac.kr} \\
% {\tt\small \{hyunsu1125.kim}@navercorp.com} \\
% {\tt\small \{gayoung.lee}@navercorp.com} \\
%{\tt\small \{sungjoo.yoo}@gmail.com} \\
% {\tt\small \{yunjey.choi}@navercorp.com} \\
}
\maketitle

%%%%%%%%% ABSTRACT
\begin{abstract}
   %GANs using implicit neural representation (inrGANs) have many special properties that convolutional GANs do not have, such as zero-shot super-resolution and out-painting, opening a new horizon for the application of GANs. However, since the networks are entirely composed of fully-connected layers, inrGANs require a lot of GPU memory for training because they do not get computational benefits from convolutional layers. Since the number of GPU required for training increases in proportion to the GPU memory used for training, it is becoming the largest bottleneck that makes it difficult to apply inrGANs in practice. To mitigate this problem, we propose multi-stage patch training method for training inrGANs with small GPU memory. We also propose a novel reconstruction loss designed to solve global coherence problem of generated images during multi-stage patch training. This allows inrGANs training using three times less GPU memory, ** times less training time while maintaining FID score at a reasonable level.
   
   Recent studies have shown remarkable progress in GANs based on implicit neural representation (INR) - an MLP that produces an RGB value given its (x, y) coordinate. They represent an image as a continuous version of the underlying 2D signal instead of a 2D array of pixels, which opens new horizons for GAN applications (e.g., zero-shot super-resolution, image outpainting). However, training existing approaches require a heavy computational cost proportional to the image resolution, since they compute an MLP operation for every (x, y) coordinate. To alleviate this issue, we propose a multi-stage patch-based training, a novel and scalable approach that can train INR-based GANs with a flexible computational cost regardless of the image resolution. Specifically, our method allows to generate and discriminate by patch to learn the local details of the image and learn global structural information by a novel reconstruction loss to enable efficient GAN training. We conduct experiments on several benchmark datasets to demonstrate that our approach enhances baseline models in GPU memory while maintaining FIDs at a reasonable level. %For the research community, we will release the code publicly.
   
\end{abstract}

%%%%%%%%% BODY TEXT
\vspace{-2mm}
\section{Introduction}
\label{sec:intro}

%Recent advances in GANs enable the creation of fake images which are difficult to distinguish from real images,
Recent advances in Generative Adversarial Networks (GANs)~\cite{goodfellow2014gan,karras2019stylegan,karras2020stylegan2} enable realistic image synthesis and show practical and diverse applicability such as image-to-image translation~\cite{isola2017image,choi2018stargan,choi2020stargan,Kim2020U-GAT-IT,kim2019tag2pix}, 3d-aware image generation~\cite{chan2021pi,niemeyer2021giraffe,or2021stylesdf,gu2021stylenerf}, real image editing~\cite{abdal2019image2stylegan,zhu2020indomaingan,kim2021exploiting}, etc. Typical GANs view images as 2D pixel arrays and build them using convolutional filters. However, thanks to the success of NeRF in 3D modeling, it is also getting popular to view images as a continuous function in GANs. Implicit Neural Representations (INR)~\cite{sitzmann2020implicit,genova2019learning,park2019deepsdf,atzmon2020sal,mildenhall2020nerf,yu2022generating} is a popular method that use a neural network to approximate the continuous function. A number of recent studies including CIPS~\cite{cips} and INR-GAN~\cite{inr_gan} have proposed a model that combines the INR concept and GANs. These INR-based GANs can naturally and easily do what was difficult in convolutional GANs, such as partial patch generation, zero-shot super-resolution, and image extrapolation.

\iffalse
 \begin{figure}[t] %%% t: top, b: bottom, h: here
\begin{center}
\includegraphics[width=1.0\linewidth]{fid-gpu.png}
\end{center}
\caption{\gy{Comparison on GPU memory and FID scores between the baselines and our method.}}
\label{fig:long}
\label{fig:onecol}
\end{figure}
\fi

%Implicit neural representation was initially designed neural network learning a 3d shape.
%However, recent studies like NeRF show various possibilities of implicit neural representation, triggering researches applying implicit neural representation to GANs.
% Seminal works are CIPS and INR-GAN, which achieve almost same image generation quality of StyleGAN2, as well as the possibility of very unique applications like zero-shot super-resolution and out-painting that convolutional GANs do not have, expanding the application scope of GANs.

Despite the advantages of INR-based GANs, it is difficult to train them because they are hardware intensive due to a lot of network inference proportional to the image size. Unlike convolutional GANs~\cite{karras2020stylegan2, lin2019cocogan} which use upsampling and convolutional filters, pure INR-based GANs need to infer each coordinate of an image, so it consumes much more GPU memory.
% have intermediate feature maps whose spatial sizes are the same as target images
 For example, CIPS requires 4 times more GPU memory than StyleGAN2. Therefore, reducing computation costs is an important research topic to practically use INR-based GANs. INR-GAN reduces the costs in the generator by factorizing the parameters and progressively growing the feature maps similar to StyleGAN2. However, their method is still computationally expensive because it starts with a feature map of large size $(64^2)$ and requires generating the entire image for the discriminator. %However, INR-based GAN training is very hardware intensive because it consists of fully-connected layers not convolutional layers which leads to consume GPU memory a lot.
%In the case of GPU memory, the more necessary it is for training, the more critical it is because the total number of GPU increases in proportion.
%In fact, CIPS require 4 times more GPU memory than StyleGAN2 and the gap of hardware requirement is a major obstacle of using CIPS.

\begin{figure*}[t] %%%
\begin{center}
\includegraphics[width=0.9\linewidth]{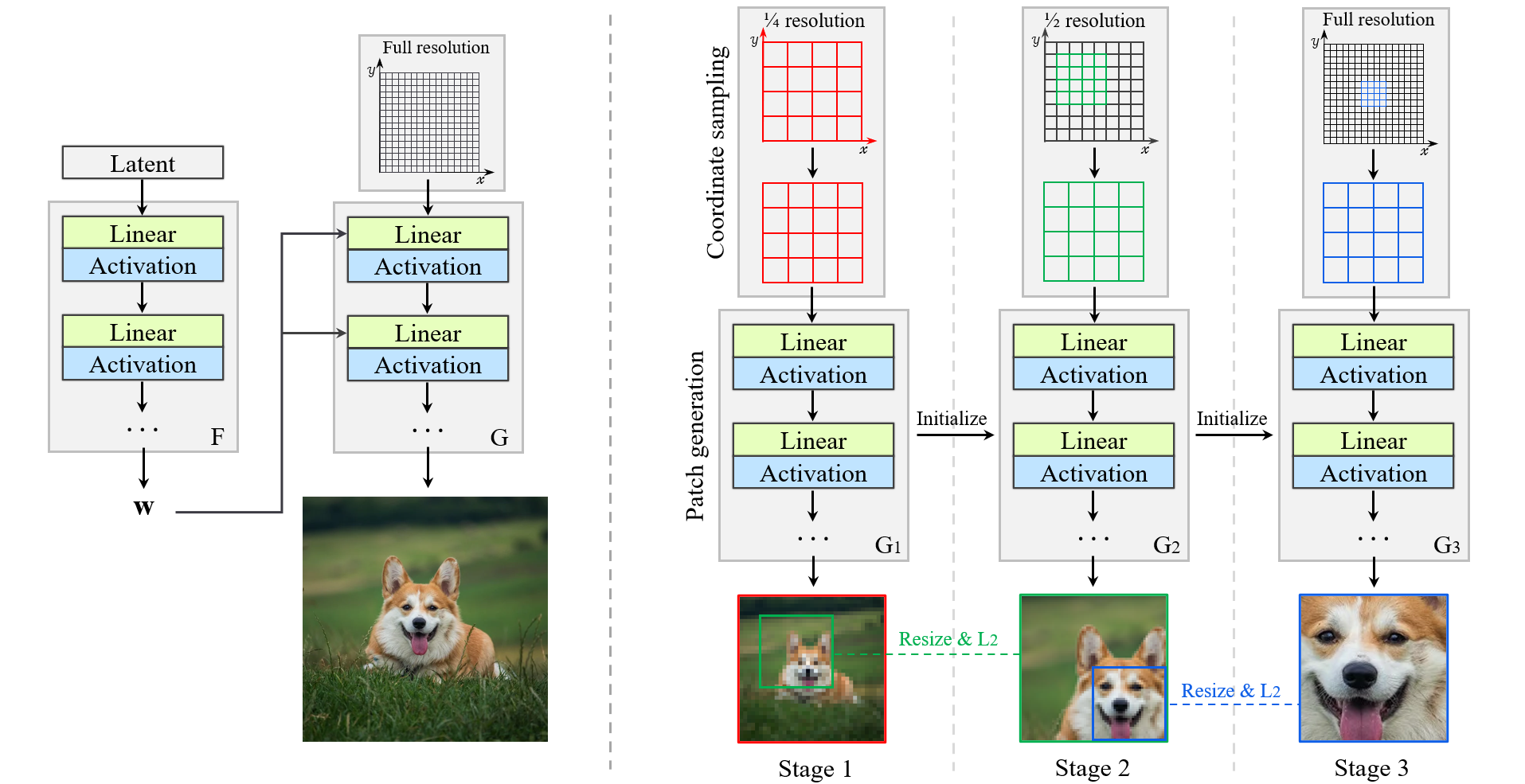}\\
\makebox[0.4\linewidth][c]{\footnotesize{(a) Traditional INR-based generator}}\hfill
\makebox[0.6\linewidth][c]{\footnotesize{(b) Multi-stage patch-based training (Ours)}}\hfill \\
\end{center}
\vspace{-3mm}
\caption{\textbf{Traditional vs. Multi-stage patch-based training.} (a) Training existing INR-based GANs \cite{inr_gan,cips} is computationally expensive as they require performing an MLP operation $G$ on all (x, y) coordinates for full resolution ($16^2$ in the example). (b) Our proposed multi-stage patch-based training enables efficient training of INR-based GANs by performing $G$ only on a predetermined small number of (x, y) coordinates ($4^2$ in the example) regardless of resolution. In the early stage (Stage 1), a coarse global image is generated from the sparse grid, and in the later stages (Stage 2,\thinspace 3), local patches with fine details are generated from the dense grids. The local patch generated in each later stage is regularized to match the corresponding region in the image generated in the previous stage. In multi-stage patch-based training, we omit the mapping network $F$ for brevity.}
\label{fig:method}
\end{figure*} %%% Figure 9.

In this paper, we propose a method that can dramatically reduce the training costs for INR-based GAN using multi-stage patch-based training. During training, our method generates small patches~($32^2$) instead of entire images, and the generated patches are fed to the discriminator. This patch-wise training can save a lot of GPU memory, but since the discriminator only sees the patches, it cannot give feedback on global structures. To solve this problem, we propose a novel multi-stage training method and progressively reduce the receptive field of each stage patch. Specifically, in the initial stage, the target patch is coarsely and globally sampled from an image, whereas in the final stage, the patch of equal size is densely and locally sampled. Then, in order to transfer the knowledge about the global structure of the previous stage to the current stage, we apply the consistency loss between the current generated patches and the patches cropped from the previously generated patches. By doing this, the final generator can generate a globally consistent image while it is trained using only local patches. We conduct extensive experiments with various datasets and show that our method reduces the required size of GPU memory and training time effectively while maintaining the quality of generated images comparable to the existing methods.
%In this paper, we propose multi-stage patch training method that can train INR-based GAN with low GPU memory.
%Through multi-stage patch training, we can train INR-based GAN using three times less GPU memory than common training methods, and compared to this dramatic reduction, FID remains reasonable.
%We can reduce training time ** times through multi-stage patch training, too. 
%This is a very encouraging result compared to the proportional increase in training time when gradient accumulation is applied to reduce GPU memory consumption.
%In addition, we propose a novel reconstruction loss that helps images generated during the multi-stage patch training maintain global consistency.
%Lastly, through empirical studies on the initialization choices of the learned constant in CIPS, we found the best  initialization strategy that maximizes the effect of multi-stage patch training for each dataset.
%We experimentally verified that the proposed method is effective through various datasets (see Fig. 1).

\section{Multi-stage patch-based training}
\label{sec:method}
We propose multi-stage patch-based training, which reduces the computational cost for training INR-based GANs. We build upon the INR-based GAN~\cite{cips} and keep every other component except the training strategy and patch regularization, including the adversarial loss, and hyperparameters. Overall framework can be shown in \Fref{fig:method}.

\iffalse
 \begin{figure}[t] %%% t: top, b: bottom, h: here
\begin{center}
\includegraphics[width=1.0\linewidth]{failure.png}
\end{center}
\caption{Generated Images of Patch Training.}
\label{fig:long}
\label{fig:onecol}
\end{figure}
\fi

For efficient training, we aim to generate local patches instead of full images (\textit{e.g.} generating $64^2$ patches instead of creating $256^2$ images can reduce the computational cost such as GPU memory by $\tfrac{1}{4}$). However, it is known that the generator $G$ cannot learn the global structure of an image by providing only small patches to the discriminator \cite{lin2019cocogan}. To alleviate this problem, we adopt multi-stage training in which the generator learns to produce a coarse full image in the early stage of training (Stage 1 in \Fref{fig:method}b) and learns to generate local patches with fine details in the later stages (Stage 2,\thinspace 3 in \Fref{fig:method}b).

\vspace{2mm}

\noindent\textbf{Sparse-to-dense coordinate sampling.} During training, we sample $(x, y)$ coordinates in a sparse-to-dense manner. We first define a set of integer pixel coordinates \texttt{grid} as:
\vspace{-1mm}
\begin{equation}
    \texttt{grid}\left(H, W, N\right) = \lbrace \left(\tfrac{H}{N}k, \tfrac{W}{N}k \right) \mid 0 \le k < N  \rbrace
\end{equation}

\noindent where $H, W$ are the height and width of training image resolution, respectively (\textit{e.g.} $256^2$), and $k$ is an integer value. A small $N$ gives sparsely sampled coordinates, while a large $N$ gives densely sampled ones. In the first stage of training, we set $N$ to $\tfrac{H}{4}$ to reduce the size of the coordinate grid to $\tfrac{1}{16}$ of its full resolution (\textit{sparse sampling}). In the second and third stages of training, we set $N$ to $\tfrac{H}{2}$ and $H$, respectively (\textit{dense sampling}). We apply appropriate random cropping to reduce the computational cost in the later stages.

\vspace{1mm}

\noindent\textbf{Coarse-to-fine patch generation.} We train the generators to produce coarse global images in the early stage of training and local patches with fine details in the later stages. Here, we denote the generator for each stage $i \in \left\{1, 2, 3\right\}$ as $G_i$ for clarity. Our generator $G_i$ takes as input a random Gaussian vector $\textbf{z} \in \mathds{R}^{128}$ shared across all pixels and pixel coordinates $\left(x, y\right) \in \left\{0\dots W-1\right\} \times \left\{0\dots H-1\right\}$. 

{
\begin{figure*}[h!]
\begin{centering}
\renewcommand{\arraystretch}{0}
\par\end{centering}
\centering
%\hfill{}%
\begin{tabular}{c@{\hskip 0.05in}c@{\hskip 0.05in}c@{\hskip 0.28in}c@{\hskip 0.05in}c@{\hskip 0.05in}c@{\hskip 0.28in}c@{\hskip 0.05in}c@{\hskip 0.05in}c@{}}
\multicolumn{3}{c}{(a) Image-based (86GB)} &\multicolumn{3}{c@{\hskip 0.28in}}{(b) Patch-based (30GB)}  & \multicolumn{3}{c}{(c) Ours (31GB)}  
\tabularnewline	
\includegraphics[width=0.095\linewidth]{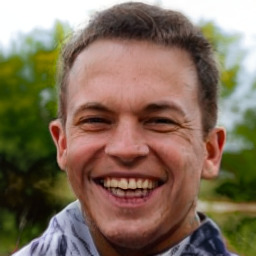} & 
\includegraphics[width=0.095\linewidth]{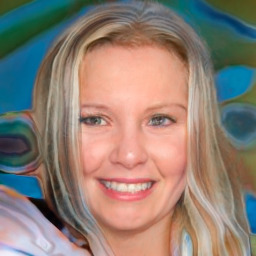} & 
\includegraphics[width=0.095\linewidth]{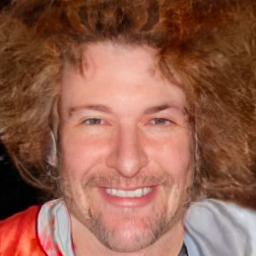} & 
\includegraphics[width=0.095\linewidth]{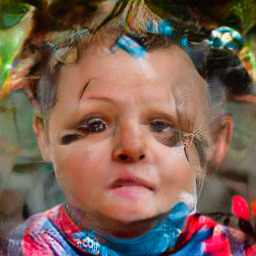} & 
\includegraphics[width=0.095\linewidth]{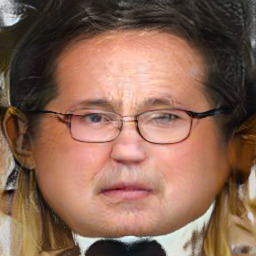} & 
\includegraphics[width=0.095\linewidth]{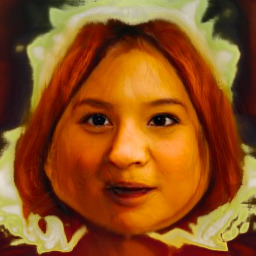} & 
\includegraphics[width=0.095\linewidth]{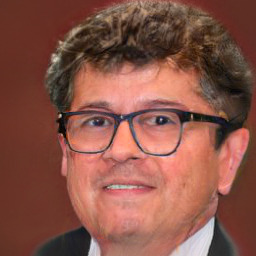} & 
\includegraphics[width=0.095\linewidth]{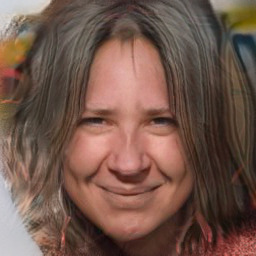} & 
\includegraphics[width=0.095\linewidth]{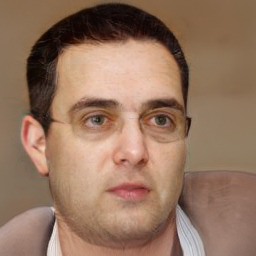} \tabularnewline 

\includegraphics[width=0.095\linewidth]{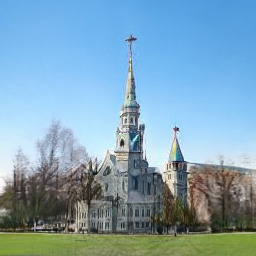} & 
\includegraphics[width=0.095\linewidth]{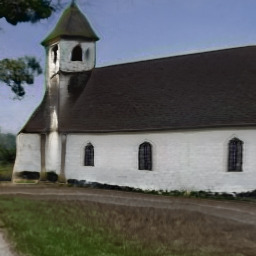} & 
\includegraphics[width=0.095\linewidth]{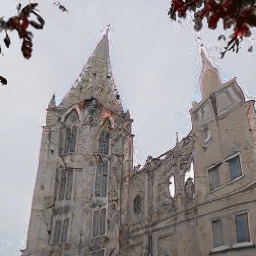} & 
\includegraphics[width=0.095\linewidth]{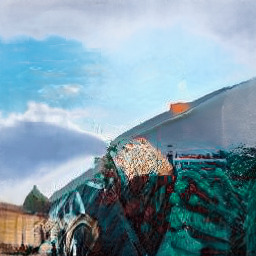} & 
\includegraphics[width=0.095\linewidth]{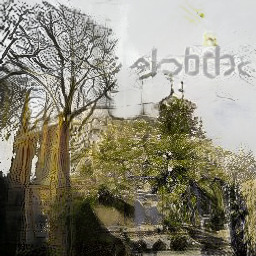} & 
\includegraphics[width=0.095\linewidth]{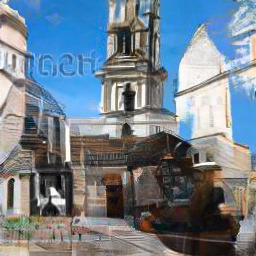} & 
\includegraphics[width=0.095\linewidth]{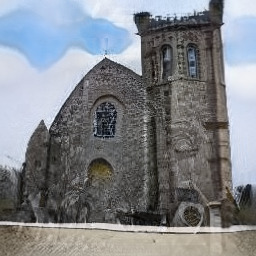} & 
\includegraphics[width=0.095\linewidth]{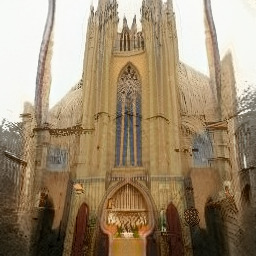} & 
\includegraphics[width=0.095\linewidth]{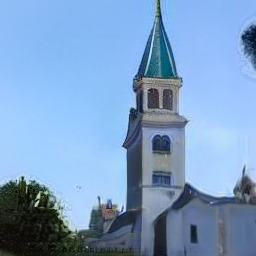} \tabularnewline 

\includegraphics[width=0.095\linewidth]{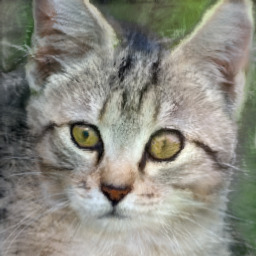} & 
\includegraphics[width=0.095\linewidth]{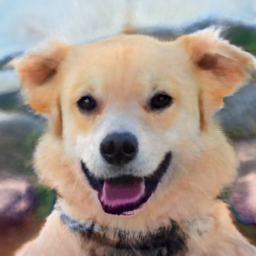} & 
\includegraphics[width=0.095\linewidth]{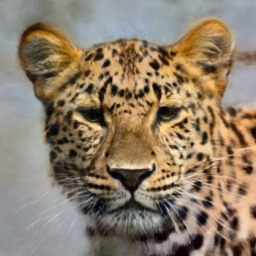} & 
\includegraphics[width=0.095\linewidth]{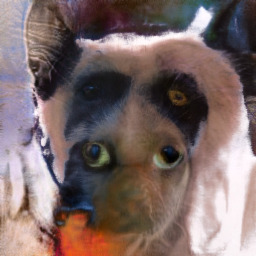} & 
\includegraphics[width=0.095\linewidth]{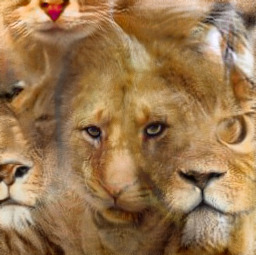} & 
\includegraphics[width=0.095\linewidth]{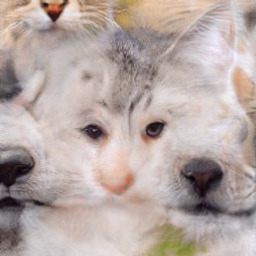} & 
\includegraphics[width=0.095\linewidth]{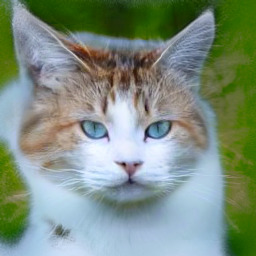} & 
\includegraphics[width=0.095\linewidth]{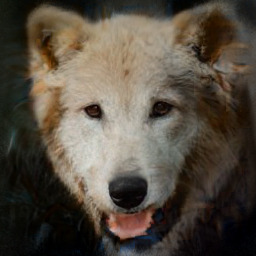} & 
\includegraphics[width=0.095\linewidth]{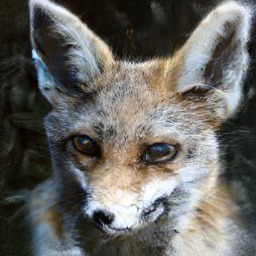}
\end{tabular}\hfill{}
\vspace{-0.5em}
\caption{\label{fig:qual_image} Qualitative comparison with the baselines and our method. The first/second/third row shows samples from FFHQ/LSUN Church/AFHQ, respectively. The image-based model offers the best quality but requires much GPU memory (86GB), whereas the patch-based model needs much less GPU memory (30GB) but generates globally inconsistent images. Our method uses the comparable amount of GPU memory (31GB) to the patch-based model, while producing much better image quality.}
\end{figure*}
}

The first stage generator $G_1$ produces a coarse global image $I_1$ by performing an MLP operation for each $\left(x, y\right)$ coordinates, while keeping random vector $\mathbf{z}$ fixed:
%In the initial stage, we first sample the coordinates $(x, y)$ sparsely: 
\begin{align}
    I_1 = \left\{ G_1\left(x, y; \mathbf{z}\right) \mid \left(x, y \right) \in \texttt{grid}\left(H, W, \tfrac{H}{4}\right) \right\}.
    \label{eq:image_gen}
\end{align}
We train $G_1$ with an adversarial loss~\cite{goodfellow2014gan} to generate images that are indistinguishable from real images of low resolution. Note that unlike traditional INR-based GANs~\cite{cips,inr_gan}, our method sets $N$ to $\tfrac{H}{4}$ instead of $H$, which efficiently reduces GPU memory.

Unlike $G_1$, we train the generators $G_2$, $G_3$ to produce local patches instead of full images. We use the generator trained in the previous stage to initialize the generator in the later stage (\textit{i.e.} initialize $G_2$ with the weights of $G_1$). This helps to distill the global representation learned in the previous stage. The equation is similar to that of $G_1$, but the (x, y) coordinates are densely sampled and randomly selected:

\vspace{-2mm}
\begin{align}
    I_i = \left\{ G_i\left(x, y; \mathbf{z}\right) \mid \left(x, y \right) \in \texttt{rcrop}(\texttt{grid}\left(H, W, N_i\right)) \right\},
    \label{eq:image_gen}
\end{align}
\noindent where \texttt{rcrop} indicates a random crop operation. We set $N_2$ to $\tfrac{H}{2}$ and $N_3$ to $H$ for $G_2$ and $G_3$, respectively. We obtain a coordinate grid of $\tfrac{1}{4}$ size compared to full resolution through the \texttt{rcrop} operation, and use the small grid to efficiently train the generators to produce local patches.

\vspace{2mm}
\noindent\textbf{Patch regularization.} In order to maintain consistency between the currently generated patch $I_i$ and the region cropped from the previously generated image (or patch) $I_{i-1}$, we apply patch regularization:

\begin{equation}
\mathcal{L}_{patch} = \mathbb{E} \left[ { \lVert \texttt{resize}(I_{i}, \tfrac{1}{2}) - \texttt{crop}(I_{i-1}) \lVert}_{2} \right],
\label{eqn:patch_reg}
\end{equation}
where $\texttt{resize}(\cdot, \tfrac{1}{2})$ reduces the size of image in half. The proposed patch regularization is simple and helps to distill the global structure learned from the previous stage to the current stage.

%-------------------------------------------------------------------------

%------------------------------------------------------------------------

% \begin{table*}[t]
% 	\setlength{\tabcolsep}{0.2em}
% 	\renewcommand{\arraystretch}{0.95}
% 	\centering
% 	\begin{tabular}{x{2cm}|x{1.8cm}|x{2.3cm}||x{1.8cm}|x{2.3cm}||x{1.8cm}|x{2.3cm}}
% 		& \multicolumn{2}{c||}{\normalsize{} FFHQ (5 days) } & \multicolumn{2}{c||}{\normalsize{} LSUN Church (6 days) }  & \multicolumn{2}{c}{\normalsize{} AFHQ (4 days)} \tabularnewline
% 		Method &   FID$\downarrow$ &  GPU mem.$\downarrow$  &  FID$\downarrow$  &  GPU mem.$\downarrow$  & FID$\downarrow$   &  GPU mem.$\downarrow$   \tabularnewline
% 		\hline 	 \hline 	
% 		{\small{} Image-based } & \small{} 8.51 & \small{}  123 & \small{} 6.42 & \small{} 123  & \small{} 10.00 & \small{} 123 \tabularnewline
% 		{\small{} Patch-based } & \small{} 41.65 & \small{} 123 & \small{} 18.48 & \small{} 123   & \small{} 39.39 & \small{} 123 \tabularnewline
% 		{\small{} Ours } & \small{} 24.38 & \small{} 123 & \small{} 10.08 & \small{} 123 & \small{} 17.13 & \small{} 123
% 		\end{tabular}
% 	\vspace{-0.5em}
% 	\caption{Comparison on FID score and used GPU memory for each method. While patch-based method is memory-efficient than the original image-based method, it produces worse quality images in terms of the FID score. Our method requires the same amount of GPU memory as the patch-based model, but produces higher quality images.}
% \label{table:comp_table} %
% \end{table*}

\begin{table*}[t]
	\setlength{\tabcolsep}{0.2em}
	\renewcommand{\arraystretch}{0.95}
	\centering
	\begin{tabular}{x{2cm}||x{2cm}|x{2cm}|x{2cm}||x{2.3cm}|x{2.3cm}}
	 & \multicolumn{3}{c||}{\normalsize{} FID Scores$\downarrow$}  &  \multicolumn{2}{c}{\normalsize{} Computation Costs} \tabularnewline
	Method & \specialcell{FFHQ\\(5 days)} & \specialcell{Church\\(6 days)} & \specialcell{AFHQ\\(4 days)} & GPU mem.$\downarrow$ & sec/iter$\downarrow$ \tabularnewline
	\hline 	 \hline 
	{\small{} Image-based } & \small{} 8.51 & \small{} 6.42 & \small{} 10.00 & \small{} 86GB & \small{} 3.04 \tabularnewline
	{\small{} Patch-based } & \small{} 41.65 & \small{} 18.48 & \small{} 39.39 & \small{} 30GB & \small{} 0.82 \tabularnewline
	{\small{} Ours } & \small{} 24.38  & \small{} 10.08 & \small{} 17.13 & \small{} 31GB	& \small{} 0.71
    \end{tabular}
	\vspace{-0.5em}
	\caption{Comparison on FID score and computational costs for each method. While patch-based method is memory-efficient than the original image-based method, it produces worse quality images in terms of the FID score. Our method requires the same amount of GPU memory as the patch-based model, but produces higher quality images. We also report the running time for each training iteration.}
\label{table:comp_table} %
\end{table*}

\section{Experiments}
\label{sec:formatting}

Our multi-stage patch-based method effectively reduces the required size of GPU memory ($2.8\times$ lower) in training. In this section, we conduct experiments on various benchmark datasets (FFHQ, LSUN Church, and AFHQ) to verify the effectiveness of our method.
%We conduct experiments using CIPS, which has the best performance in INR-based GANs.
% We evaluate on various benchmark datasets (FFHQ, LSUN CHURCH, AFHQ) for verifying our method produces consistent results in many cases.
All experiments are conducted at $256\times256$ scale with the $G_3$ generator, and we use the Fr\'echet inception distance (FID) metric to show that our method still retains comparable performance in image generation.
%but there is no doubt that our method can be applied to various resolutions.

% \begin{table}
% \centering
% \caption{Training Time Comparison}
% \label{t3}
% \begin{tabular}{c|ccc}
% \noalign{\smallskip}\noalign{\smallskip}\hline\hline
% & FFHQ & CHURCH & AFHQ \\
% \hline
% Patch & 6d 7h & 4d 21h & 4d 2h \\
% Image & 9d 11h + & 9d 20h + & 3d 15h \\
% Accumulate Gradient & ?d ?h & ?d ?h & ?d ?h \\
% Ours & 5d 7h + & 5d 10h + & 7d 20h \\
% \hline
% \hline
% \end{tabular}
% \end{table}

%-------------------------------------------------------------------------
\subsection{Baseline Models}

Since CIPS~\cite{cips} is one of the state-of-the-art INR-based GANs, we demonstrates the applicability of our method to the CIPS model. To show the effectiveness of our method, we compare our method with three baselines.

\noindent\textbf{Image-based} method is the original version of CIPS network. We do not change any configurations from its paper. 

\noindent\textbf{Patch-based} method is the patch-based training without our multi-stage training and patch regularization. The network is trained with $4\times$ smaller patches and only adversarial loss term.

\noindent\textbf{Gradient Accumulation} is the same as \textbf{Image-based} method except for the batch size. To avoid the GPU memory limitation, some recent works~\cite{karras2019stylegan,karras2020stylegan2} may use small batch size and accumulate gradients. The network weights are updated once every multiple batches, whose summation is equal to that of the original batch size.

%-------------------------------------------------------------------------
% \subsection{Qualitative results}

% In order to obtain a qualitative evaluation of our method, we compare our approach to image training, which consumes high GPU memory and patch training, which consumes low GPU memory.
% As see in Fig. 4, our method creates realistic images whose quality is close to image training.
% In addition, structures of images generated from patch training are unrecognizable, but we alleviate this problem through our novel reconstruction loss.
% Considering that our method shows competitive results on all benchmark datasets, our method can be adopted on various situations need INR-based GANs.

%-------------------------------------------------------------------------

\subsection{Main results}
\Fref{fig:qual_image} and \Tref{table:comp_table} show the qualitative and quantitative results. For a fair comparison, we trained all baselines with the same training time; 4, 5, 6 days for AFHQ, FFHQ, and LSUN Church, respectively. We set the training time in proportion to the size of the data. Gradient Accumulation method is excluded from \Tref{table:comp_table} because it needs $n \times$ more time if we want to use $n \times$ smaller batch size.  Ours shows visually comparable quality compared to the original CIPS network while it needs $2.8\times$ less memory of GPU. Without our multi-stage training, image quality deteriorates significantly in the patch-based method. Our method needs only $3\%$ additional GPU memory but shows significantly better image generation quality than patch-based method according to the FID score; FIDs increase by 17.27, 8.40, and 22.26 in FFHQ, LSUN Church, and AFHQ, respectively. Since our method and the patch-based model generate only part of an image, each training iteration takes significantly less time than the image-based model, and we can run more training iterations in the same amount of time. Note that our method is slightly faster than the patch-based model because we can skip random cropping for the first stage.

\begin{figure}[t] %%% t: top, b: bottom, h: here
\centering
\begin{tabular}{c@{\hskip 0.1in}c@{\hskip 0.1in}c}
\includegraphics[width=0.25\linewidth]{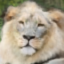} &
\includegraphics[width=0.25\linewidth]{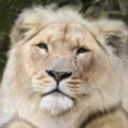} &
\includegraphics[width=0.25\linewidth]{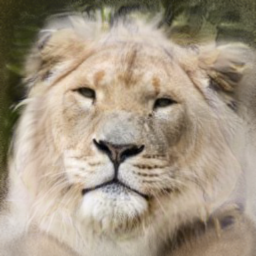} \\
\includegraphics[width=0.25\linewidth]{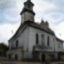} &
\includegraphics[width=0.25\linewidth]{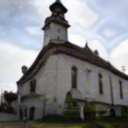} &
\includegraphics[width=0.25\linewidth]{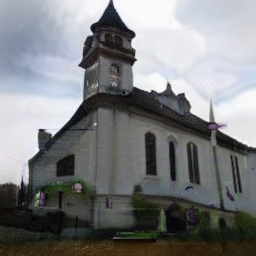} \\
Stage 1 & Stage 2 & stage 3 \\
\end{tabular}
\caption{Samples of each training phase in our multi-stage training method. In the first stage, coarse and global contours are generated, and in the later stage, more and more details are added. The ability to produce globally consistent images is transferred by our patch regularization loss.}
\label{fig:long}
\label{fig:onecol}
\end{figure}

%-------------------------------------------------------------------------
% \subsection{FID - Memory}

% As see in Figure 1, image training shows high GPU memory with low FID score, otherwise patch training shows low GPU memory with high FID score.
% Our method drastically reduces GPU memory while minimizing performance degradation, therefore shows low GPU memory with competitive FID score.

%-------------------------------------------------------------------------
% \subsection{Training time comparison}

% As see in Table 1, our method needs 50\% of training time compared to image training, while produces competitive results.
% Gradient accumulation can be applied for reducing GPU memory, however it takes longer time choosing one from time - memory trade-off.
% Exceptionally, our method on small dataset (AFHQ) reports worse result.
% Judging from the deterioration in patch training time, image training seems to converge very quickly because of the small data size.
% However, in reality, handling large-scale datasets is promising and mandatory direction of GAN's future, and our method is useful because it shows significant time improvement in large-scale datasets (FFHQ, LSUN CHURCH).

%-------------------------------------------------------------------------
\subsection{Effect of patch regularization}

In multi-stage patch-based training, we propose patch regularization which matches the generated patches of different training phases as we've discussed in \Sref{sec:method}. \Fref{fig:onecol} shows our regularization makes the network produce consistent structure in all stages. Stage 1 shows blurry but structurally meaningful images, and stage 3 shows high-fidelity images while maintaining the structure of the early stages. Without this loss term, our network cannot fully exploit the advantage of the multi-stage training.

\begin{figure}[t] %%% t: top, b: bottom, h: here
\centering
\begin{tabular}{@{\hskip 0in}c@{\hskip 0.05in}c}
\includegraphics[width=0.45\linewidth]{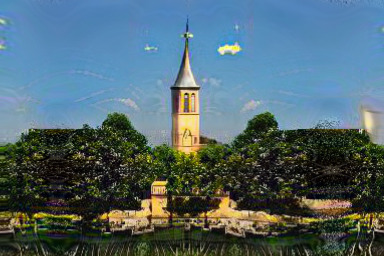} &
\includegraphics[width=0.45\linewidth]{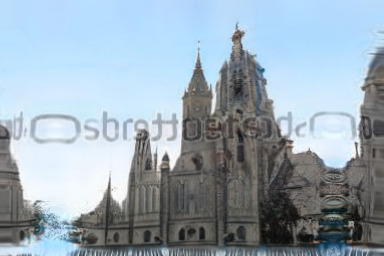} \\
\end{tabular}
\caption{Extrapolation on LSUN Church using our method. The pixels in out-of-boundary locations are properly generated.}
\label{fig:long}
\label{fig:extrapol}
\end{figure}

\subsection{Extrapolation Results}
In \Fref{fig:extrapol}, we show the results of extrapolation on LSUN Church using our method. Thanks to the advantages of INR-based model, our method can generate an image of a size not seen during training by simply feeding the targeted coordinates.
% Reconstruction loss really helps generator to learn structural information, as illustrated in Fig. 5.
% As the training stage progresses, images are further improved with high-resolution details, but overall structures are still maintained.

%------------------------------------------------------------------------
\section{Conclusion and Discussion}
In this paper, we propose multi-stage patch-based training, a novel and scalable approach that can train INR-based GANs with a flexible computational cost regardless of the image resolution. We conducted experiments on several benchmark datasets and demonstrated that our method contributes to reducing the required size of GPU memory in training INR-based GAN models. 

Our method also has some limitations. The proposed patch regularization might be too restrictive as it forces the patch generated in the current stage to be strongly match the image in the previous step. Also, the performance of multi-stage training for a specific dataset (i.e. FFHQ) could be more improved. Improving the performance and devising more flexible regularization to extract global structures would be one of the meaningful future work.

\noindent\textbf{Acknowledgements.} The authors thank NAVER AI Lab researchers for constructive discussion. All experiments were conducted on NAVER Smart Machine Learning (NSML) platform~\cite{kim2018nsml, sung2017nsml}.

% inference는 여전히 느림(?, 이게 cips 같은 경우에는 오히려 속도는 조금 빨라서 쓰기 어려울 것 같습니다), 특정 데이터셋에 대해서 성능 개선이 덜됨. 이는 future work로.
% global 정보를 배우기 위해 걸어주는 mse loss가 사실 global 정보를 직접적으로 배우는건 아니고, reconstruction으로 강제하는 것 뿐임
% inr을 3d-gan에 많이 쓰는데 3d-gan은 이거보다 더 메모리 많이 먹을테니, 3d 분야에서 응융하는 Future work 기대?
% inr은 fc layer로 이루어져 있는데, convolutional gan에서 사용하는 Modulation과 동일한 방식을 채용하는데 inr에 더 최적화된 modulation이 있을지도?

%Q) 해상도는 1/4씩 줄여 실제 1/16크기로 줄였는데, memory는 1/3정도 밖에 줄지 않은 이유? 여러개의 g, d를 써서 그런건가요?
% 아뇨 애초에 1/4 정도로 줄고, multi-gpu overhead가 좀 있어서 1/3 수준으로 된 것 같습니다. 여러개의 g,d 때문에 오른것은 1GB 정도로 거의 차지하지 않는 것 같아요. patch=30GB, ours=31GB

%이것도 오해의 여지가 있는데 설명좀 해주면 좋을것 같긴하네요

%1/4보다 더 줄이면 안되는거죠? 
% 아 이거는 실험해볼 생각을 못했어서 (애초에 1/4에서도 image랑 동일한 성능을 못내서, image랑 동일한 성능을 낼수있는 방안들에 대해서만 계속 고민을하고 더 작은데서 실험하는 것은 생각을 못했습니다) 결과를 잘 모르겠습니다.
% 근데 아래에 내용 쓰고나니까 잘 안될거 같다는 생각이 들긴 하네요 패치 작아질수록 structure 정보 없어지게되니

%ffhq에서 성능개선정도가 아쉬운 이유가 뭘까요? 짐작가는 원인? 이게 limitation에서 설명하는게 적합한것 같아요.
% 머리카락이 상당히 큰 비율을 차지하는 경우 (특히 여자), 머리카락 같은 경우가 사실 strucure 정보가 사실상 없는거라서, 그게 structure 정보가 조금은 있어야 얘도 continuous하게 이어지도록 reconstruction으로 학습을 하는데, 머리카락이 이미지의 많은 비율을 차지하고있으면 그게 좀 어려운 것 같습니다.
% 그래서 남자는 잘 생성되는데, 여자가 잘 안되는 거 같기도 하고요
% 어차피 ffhq에서 숫자도 안좋은거 눈에보이고 limitation에서 이유도 설명할거면 잘안되는거 이미지도 그냥 보여줄까요 차라리, ffhq에서 특히 여자가 엄청 특이한 케이스고 남자는 또 잘되니까 나쁘지는 않을 것 같기도 합니다.
% ffhq가 다 안되는게아니라 여자만 잘 안되서 성능이 좀 안좋다라는 느낌으로?

%%%%%%%%% REFERENCES
{\small
\bibliographystyle{ieee_fullname}
\bibliography{egbib}

\begin{thebibliography}{10}\itemsep=-1pt

\bibitem{abdal2019image2stylegan}
Rameen Abdal, Yipeng Qin, and Peter Wonka.
\newblock Image2stylegan: How to embed images into the stylegan latent space?
\newblock In {\em CVPR}, 2019.

\bibitem{cips}
Ivan Anokhin, Kirill Demochkin, Taras Khakhulin, Gleb Sterkin, Victor
  Lempitsky, and Denis Korzhenkov.
\newblock Image generators with conditionally-independent pixel synthesis.
\newblock {\em arXiv preprint arXiv:2011.13775}, 2020.

\bibitem{atzmon2020sal}
Matan Atzmon and Yaron Lipman.
\newblock Sal: Sign agnostic learning of shapes from raw data.
\newblock In {\em CVPR}, 2020.

\bibitem{chan2021pi}
Eric~R Chan, Marco Monteiro, Petr Kellnhofer, Jiajun Wu, and Gordon Wetzstein.
\newblock pi-gan: Periodic implicit generative adversarial networks for
  3d-aware image synthesis.
\newblock In {\em CVPR}, 2021.

\bibitem{choi2018stargan}
Yunjey Choi, Minje Choi, Munyoung Kim, Jung-Woo Ha, Sunghun Kim, and Jaegul
  Choo.
\newblock Stargan: Unified generative adversarial networks for multi-domain
  image-to-image translation.
\newblock In {\em CVPR}, 2018.

\bibitem{choi2020stargan}
Yunjey Choi, Youngjung Uh, Jaejun Yoo, and Jung-Woo Ha.
\newblock Stargan v2: Diverse image synthesis for multiple domains.
\newblock In {\em CVPR}, 2020.

\bibitem{genova2019learning}
Kyle Genova, Forrester Cole, Daniel Vlasic, Aaron Sarna, William~T Freeman, and
  Thomas Funkhouser.
\newblock Learning shape templates with structured implicit functions.
\newblock In {\em ICCV}, 2019.

\bibitem{goodfellow2014gan}
Ian Goodfellow, Jean Pouget-Abadie, Mehdi Mirza, Bing Xu, David Warde-Farley,
  Sherjil Ozair, Aaron Courville, and Yoshua Bengio.
\newblock Generative adversarial networks.
\newblock In {\em NeurIPS}, 2014.

\bibitem{gu2021stylenerf}
Jiatao Gu, Lingjie Liu, Peng Wang, and Christian Theobalt.
\newblock Stylenerf: A style-based 3d-aware generator for high-resolution image
  synthesis.
\newblock {\em ICLR}, 2022.

\bibitem{isola2017image}
Phillip Isola, Jun-Yan Zhu, Tinghui Zhou, and Alexei~A Efros.
\newblock Image-to-image translation with conditional adversarial networks.
\newblock In {\em CVPR}, 2017.

\bibitem{karras2019stylegan}
Tero Karras, Samuli Laine, and Timo Aila.
\newblock A style-based generator architecture for generative adversarial
  networks.
\newblock In {\em CVPR}, 2019.

\bibitem{karras2020stylegan2}
Tero Karras, Samuli Laine, Miika Aittala, Janne Hellsten, Jaakko Lehtinen, and
  Timo Aila.
\newblock Analyzing and improving the image quality of stylegan.
\newblock In {\em CVPR}, 2020.

\bibitem{kim2021exploiting}
Hyunsu Kim, Yunjey Choi, Junho Kim, Sungjoo Yoo, and Youngjung Uh.
\newblock Exploiting spatial dimensions of latent in gan for real-time image
  editing.
\newblock In {\em CVPR}, 2021.

\bibitem{kim2019tag2pix}
Hyunsu Kim, Ho~Young Jhoo, Eunhyeok Park, and Sungjoo Yoo.
\newblock Tag2pix: Line art colorization using text tag with secat and changing
  loss.
\newblock In {\em ICCV}, 2019.

\bibitem{kim2018nsml}
Hanjoo Kim, Minkyu Kim, Dongjoo Seo, Jinwoong Kim, Heungseok Park, Soeun Park,
  Hyunwoo Jo, KyungHyun Kim, Youngil Yang, Youngkwan Kim, et~al.
\newblock Nsml: Meet the mlaas platform with a real-world case study.
\newblock {\em arXiv preprint arXiv:1810.09957}, 2018.

\bibitem{Kim2020U-GAT-IT}
Junho Kim, Minjae Kim, Hyeonwoo Kang, and Kwang~Hee Lee.
\newblock U-gat-it: Unsupervised generative attentional networks with adaptive
  layer-instance normalization for image-to-image translation.
\newblock In {\em ICLR}, 2020.

\bibitem{lin2019cocogan}
Chieh~Hubert Lin, Chia{-}Che Chang, Yu{-}Sheng Chen, Da{-}Cheng Juan, Wei Wei,
  and Hwann{-}Tzong Chen.
\newblock {COCO-GAN:} generation by parts via conditional coordinating.
\newblock In {\em ICCV}, 2019.

\bibitem{mildenhall2020nerf}
Ben Mildenhall, Pratul~P Srinivasan, Matthew Tancik, Jonathan~T Barron, Ravi
  Ramamoorthi, and Ren Ng.
\newblock Nerf: Representing scenes as neural radiance fields for view
  synthesis.
\newblock In {\em ECCV}, 2020.

\bibitem{niemeyer2021giraffe}
Michael Niemeyer and Andreas Geiger.
\newblock Giraffe: Representing scenes as compositional generative neural
  feature fields.
\newblock In {\em CVPR}, 2021.

\bibitem{or2021stylesdf}
Roy Or-El, Xuan Luo, Mengyi Shan, Eli Shechtman, Jeong~Joon Park, and Ira
  Kemelmacher-Shlizerman.
\newblock Stylesdf: High-resolution 3d-consistent image and geometry
  generation.
\newblock 2022.

\bibitem{park2019deepsdf}
Jeong~Joon Park, Peter Florence, Julian Straub, Richard Newcombe, and Steven
  Lovegrove.
\newblock Deepsdf: Learning continuous signed distance functions for shape
  representation.
\newblock In {\em CVPR}, 2019.

\bibitem{sitzmann2020implicit}
Vincent Sitzmann, Julien Martel, Alexander Bergman, David Lindell, and Gordon
  Wetzstein.
\newblock Implicit neural representations with periodic activation functions.
\newblock 2020.

\bibitem{inr_gan}
Ivan Skorokhodov, Savva Ignatyev, and Mohamed Elhoseiny.
\newblock Adversarial generation of continuous images.
\newblock {\em CVPR}, 2020.

\bibitem{sung2017nsml}
Nako Sung, Minkyu Kim, Hyunwoo Jo, Youngil Yang, Jingwoong Kim, Leonard Lausen,
  Youngkwan Kim, Gayoung Lee, Donghyun Kwak, Jung-Woo Ha, et~al.
\newblock Nsml: A machine learning platform that enables you to focus on your
  models.
\newblock {\em arXiv preprint arXiv:1712.05902}, 2017.

\bibitem{yu2022generating}
Sihyun Yu, Jihoon Tack, Sangwoo Mo, Hyunsu Kim, Junho Kim, Jung-Woo Ha, and
  Jinwoo Shin.
\newblock Generating videos with dynamics-aware implicit generative adversarial
  networks.
\newblock 2022.

\bibitem{zhu2020indomaingan}
Jiapeng Zhu, Yujun Shen, Deli Zhao, and Bolei Zhou.
\newblock In-domain gan inversion for real image editing.
\newblock In {\em ECCV}, 2020.

\end{thebibliography}
}

\end{document}